\documentclass[sn-mathphys-num]{sn-jnl}

\usepackage{graphicx}%
\usepackage{multirow}%
\usepackage{amsmath,amssymb,amsfonts}%
\usepackage{amsthm}%
\usepackage{mathrsfs}%
\usepackage[title]{appendix}%
\usepackage{xcolor}%
\usepackage{textcomp}%
\usepackage{manyfoot}%
\usepackage{booktabs}%
\usepackage{tabularx}
\usepackage{algorithm}%
\usepackage{algorithmicx}%
\usepackage{algpseudocode}%
\usepackage{listings}%
\theoremstyle{thmstyleone}%
\theoremstyle{thmstyletwo}%

\theoremstyle{thmstylethree}%

\begin{document}

\title{Towards the Automatic Segmentation, Modeling and Meshing of the Aortic Vessel Tree from Multicenter Acquisitions: An Overview of the SEG.A. 2023 Segmentation of the Aorta Challenge }


\author*[1,2,3]{\fnm{Yuan} \sur{Jin}}\email{jin@student.tugraz.at}
\author*[3]{\fnm{Antonio} \sur{Pepe}}\email{antonio.pepe@student.tugraz.at}
\author[4]{\fnm{Gian Marco} \sur{Melito}}
\author[5]{\fnm{Yuxuan} \sur{Chen}}
\author[6]{\fnm{Yunsu} \sur{Byeon}}
\author[6,7]{\fnm{Hyeseong} \sur{Kim}}
\author[6]{\fnm{Kyungwon} \sur{Kim}}
\author[6]{\fnm{Doohyun} \sur{Park}}
\author[8]{\fnm{Euijoon} \sur{Choi}}
\author[6,7,9,10]{\fnm{Dosik} \sur{Hwang}}
\author[11]{\fnm{Andriy} \sur{Myronenko}}
\author[11]{\fnm{Dong} \sur{Yang}}
\author[11]{\fnm{Yufan} \sur{He}}
\author[11]{\fnm{Daguang} \sur{Xu}}
\author[12]{\fnm{Ayman} \sur{El-Ghotni}}
\author[12]{\fnm{Mohamed} \sur{Nabil}}
\author[12]{\fnm{Hossam} \sur{El-Kady}}
\author[12]{\fnm{Ahmed} \sur{Ayyad}}
\author[12]{\fnm{Amr} \sur{Nasr}}
\author[13,14]{\fnm{Marek} \sur{Wodzinski}}
\author[14,15]{\fnm{Henning} \sur{Müller}}
\author[6]{\fnm{Hyeongyu } \sur{Kim}}
\author[6,16]{\fnm{Yejee } \sur{Shin}}
\author[17]{\fnm{Abbas} \sur{Khan}}
\author[17]{\fnm{Muhammad} \sur{Asad}}
\author[17, 18]{\fnm{Alexander} \sur{Zolotarev}}
\author[17, 18]{\fnm{Caroline} \sur{Roney}}
\author[19]{\fnm{Anthony} \sur{Mathur}}
\author[20]{\fnm{Martin} \sur{Benning}}
\author[17]{\fnm{Gregory} \sur{Slabaugh}}
\author[21]{\fnm{Theodoros Panagiotis} \sur{Vagenas}}
\author[21]{\fnm{Konstantinos} \sur{Georgas}}
\author[21]{\fnm{George K.} \sur{Matsopoulos}}
\author[22]{\fnm{Jihan} \sur{Zhang}}
\author[23]{\fnm{Zhen} \sur{Zhang}}
\author[23]{\fnm{Liqin} \sur{Huang}}
\author[24]{\fnm{Christian Mayer}}
\author[24]{\fnm{Heinrich} \sur{Mächler}}
\author*[1,3,25,26,27]{\fnm{Jan} \sur{Egger}}\email{jan.egger@uk-essen.de}


\affil[1]{\orgdiv{Institute for Artificial Intelligence (AI) in Medicine (IKIM)}, \orgname{University Medicine Essen (AöR)}, \orgaddress{\street{Girardetstr. 2}, \city{Essen}, \postcode{45131}, \state{NRW}, \country{Germany}}}

\affil[2]{\orgdiv{Research Center for Frontier Fundamental Studies}, \orgname{ZhejiangLab}, \orgaddress{\city{Hangzhou}, \postcode{311121}, \state{Zhejiang}, \country{China}}}

\affil[3]{\orgdiv{Institute of Computer Graphics and Vision (ICG)}, \orgname{Graz University of Technology}, \orgaddress{\street{Inffeldgasse 16/II}, \city{Graz}, \postcode{8010}, \state{Styria}, \country{Austria}}}

\affil[4]{\orgdiv{Institute of Mechanics (IFM)}, \orgname{Graz University of Technology}, \orgaddress{\street{Kopernikusgasse 24/IV}, \city{Graz}, \postcode{8010}, \state{Styria}, \country{Austria}}}

\affil[5]{\orgdiv{Zhejiang Laboratory of Philosophy and Social Sciences - Laboratory of Intelligent Society and Governance}, \orgname{ZhejiangLab}, \orgaddress{\city{Hangzhou}, \postcode{311121}, \state{Zhejiang}, \country{China}}}

\affil[6]{\orgdiv{School of Electrical and Electronic Engineering}, \orgname{Yonsei University}, \orgaddress{\city{Seoul}, \country{Republic of Korea}}}

\affil[7]{\orgdiv{Artificial Intelligence and Robotics Institute}, \orgname{Korea Institute of Science and Technology}, \orgaddress{\city{Seoul}, \country{Republic of Korea}}}

\affil[8]{\orgdiv{Department of Artificial Intelligence}, \orgname{Yonsei University}, \orgaddress{\city{Seoul}, \country{Republic of Korea}}}

\affil[9]{\orgdiv{Department of Oral and Maxillofacial Radiology}, \orgname{Yonsei University College of Dentistry}, \orgaddress{\city{Seoul}, \country{Republic of Korea}}}

\affil[10]{\orgdiv{Department of Radiology and Research Institute of Radiological Science and Center for Clinical Imaging Data Science}, \orgname{Yonsei University College of Medicine}, \orgaddress{\city{Seoul}, \country{Republic of Korea}}}
\affil[11]{\orgdiv{NVIDIA}, \orgaddress{\city{Santa Clara}, \state{CA}, \country{United States}}}
\affil[12]{\orgdiv{Brightskies Inc.}, \orgaddress{\city{Alexandria}, \country{Egypt}}}
\affil[13]{\orgdiv{Department of Measurement and Electronics}, \orgname{AGH University of Krakow}, \orgaddress{\city{Krakow}, \country{Poland}}}

\affil[14]{\orgdiv{Institute of Informatics}, \orgname{University of Applied Sciences Western Switzerland}, \orgaddress{\city{Sierre}, \country{Switzerland}}}

\affil[15]{\orgdiv{Medical Faculty}, \orgname{University of Geneva}, \orgaddress{\city{Geneva}, \country{Switzerland}}}
\affil[16]{\orgname{Probe Medical}, \orgaddress{\city{Seoul}, \country{Republic of Korea}}}
\affil[17]{\orgdiv{Digital Environment Research Institute (DERI)}, \orgname{Queen Mary University of London}, \orgaddress{\country{UK}}}

\affil[18]{\orgdiv{School of Engineering and Materials Science}, \orgname{Queen Mary University of London}, \orgaddress{\country{UK}}}

\affil[19]{\orgdiv{William Harvey Research Institute (WHRI)}, \orgname{Queen Mary University of London}, \orgaddress{\country{UK}}}

\affil[20]{\orgdiv{Department of Computer Science}, \orgname{University College London}, \orgaddress{\country{UK}}}
\affil[21]{\orgdiv{Biomedical Engineering Lab (BEL), School of Electrical and Computer Engineering }, \orgname{National Technical University of Athens}, \orgaddress{\city{Athens}, \postcode{15780}, \country{Greece}}}
\affil[22]{\orgdiv{College of Computer and Data Science}, \orgname{Fuzhou University}, \orgaddress{\city{Fuzhou}, \postcode{350108}, \state{Fujian}, \country{China}}}

\affil[23]{\orgdiv{Intelligent Image Processing and Analysis Laboratory}, \orgname{Fuzhou University}, \orgaddress{\city{Fuzhou}, \postcode{350108}, \state{Fujian}, \country{China}}}

\affil[24]{\orgdiv{Division of Cardiac Surgery, Department of Surgery}, \orgname{Medical University of Grazy}, \orgaddress{\city{Graz}, \postcode{8036}, \state{Styria}, \country{Austria}}}

\affil[25]{\orgdiv{Cancer Research Center Cologne Essen (CCCE)}, \orgname{University Medicine Essen (AöR)}, \orgaddress{\street{Hufelandstr. 55}, \city{Essen}, \postcode{45147}, \state{NRW}, \country{Germany}}}

\affil[26]{\orgdiv{Center for Virtual and Extended Reality in Medicine (ZvRM)}, \orgname{University Medicine Essen (AöR)}, \orgaddress{\street{Hufelandstr. 55}, \city{Essen}, \postcode{45147}, \state{NRW}, \country{Germany}}}

\affil[27]{\orgdiv{Faculty of Computer Science}, \orgname{University of Duisburg-Essen (UDE)}, \orgaddress{\street{Schuetzenbahn 70}, \city{Essen}, \postcode{45127}, \state{NRW}, \country{Germany}}}


\abstract{The automated analysis of the aortic vessel tree (AVT) from computed tomography angiography (CTA) holds immense clinical potential, but its development has been impeded by a lack of shared, high-quality data. We launched the SEG.A. challenge to catalyze progress in this field by introducing a large, publicly available, multi-institutional dataset for AVT segmentation. The challenge benchmarked automated algorithms on a hidden test set, with subsequent optional tasks in surface meshing for computational simulations. Our findings reveal a clear convergence on deep learning methodologies, with 3D U-Net architectures dominating the top submissions. A key result was that an ensemble of the highest-ranking algorithms significantly outperformed individual models, highlighting the benefits of model fusion. Performance was strongly linked to algorithmic design, particularly the use of customized post-processing steps, and the characteristics of the training data. This initiative not only establishes a new performance benchmark but also provides a lasting resource to drive future innovation toward robust, clinically translatable tools.
}


\maketitle

\section{Introduction}\label{sec1}
The aortic vessel tree (AVT), encompassing the aorta and its branching network, serves as the primary conduit for systemic blood supply. Pathologies such as aortic aneurysms and dissections pose critical risks of fatal rupture, frequently requiring invasive surgical intervention~\cite{bossone2021epidemiology}. To monitor disease progression, patients undergo serial medical imaging examinations, where geometric comparison of AVT structures across longitudinal scans enables precise tracking of morphological changes in both the main aortic lumen and its branches.

Quantitative AVT analysis necessitates accurate segmentation of medical images and intensive manual annotation, which imposes substantial time and financial burdens. While clinical applications of AVT segmentation span surgical planning, hemodynamic modeling, and disease surveillance, reliance on manual delineation restricts its adoption in routine practice. Automated segmentation methods could bridge this gap by integrating quantitative analysis into clinical workflows, thereby enhancing diagnostic precision and therapeutic decision-making~\cite{jin2021ai}. Notably, reconstructed 3D AVT meshes further enable computational fluid dynamics simulations to elucidate hemodynamic patterns and pathological mechanisms underlying aortic diseases~\cite{song2023systematic}.

Existing segmentation approaches range from conventional threshold-based techniques to advanced deep learning architectures and hybrid methods~\cite{jin2021ai}. Although these studies validate the technical viability of AVT segmentation, methodological reproducibility and comparative evaluation remain constrained by proprietary datasets and algorithms. Moreover, predominant reliance on single-center data limits transferability across diverse clinical environments.

Medical image analysis challenges have emerged as pivotal drivers of innovation, exemplified by initiatives like the AutoImplant~\cite{li2023towards}, Medical Segmentation Decathlon~\cite{antonelli2022medical}, and autoPET challenges~\cite{gatidis2024results}. These platforms accelerate progress through task standardization, benchmark establishment, and open-access annotated datasets, fostering transparent algorithm development and validation.

To address current limitations in AVT segmentation and propel the field forward, we organized the SEG.A. challenge under the MICCAI 2023 framework. This initiative pursued three objectives: (1) catalyzing research in automated AVT segmentation, (2) establishing a standardized platform for method comparison, and (3) benchmarking state-of-the-art performance. The challenge focused on multicenter CTA-based AVT segmentation, supported by a curated dataset of 56 CTA scans with expert-validated manual annotations. We also systematically analyzed how the composition and scale of training data impact algorithm performance, yielding critical insights for the field.

\section{Results}\label{sec2}
The challenge framework was implemented in accordance with the Biomedical Image Analysis Challenges reporting guidelines~\cite{maier2020bias}. The training dataset, publicly accessible through \citet{Radl2022avtData}, comprised multi-institutional cases to ensure anatomical diversity. For unbiased performance assessment, a held-out test set was obtained from an independent institution, maintaining complete separation from training and validation data to prevent information leakage.

\subsection{Challenge participation}
\begin{figure}[h]
    \centering
    \includegraphics[width = \linewidth]{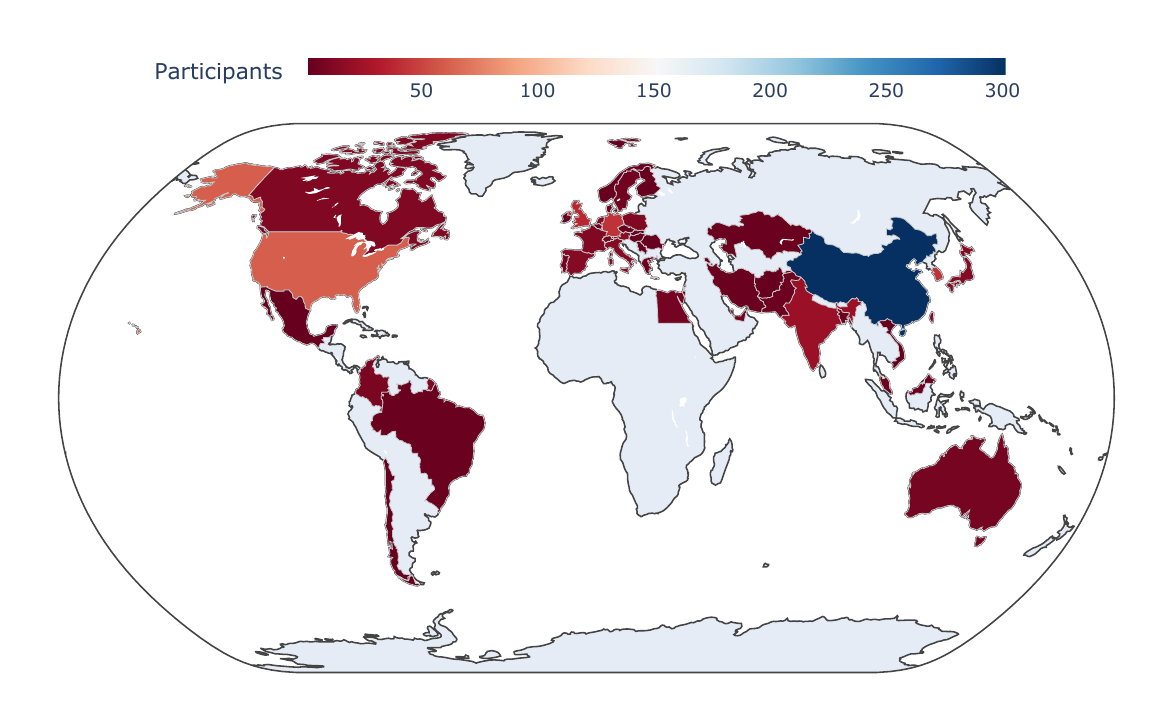}\\
    \includegraphics[width=1.0\textwidth]{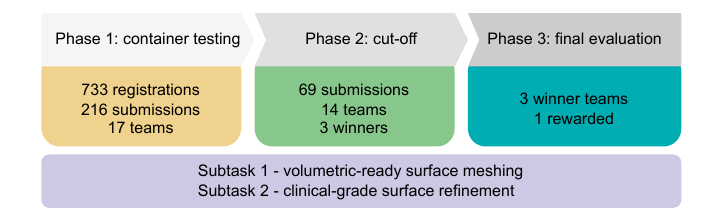}
    \caption{Overview of the challenge phases and the geographic distribution of participants. The SEG.A.challenge was structured into three main phases: Phase 1 – participants tested their Docker images and algorithms using two hidden cases; Phase 2 – five hidden cases were used for evaluation; Phase 3 – the top three teams from Phase 2 were given the opportunity to resubmit their final algorithms, with the ultimate winner receiving a monetary prize. Additionally, two optional subtasks, surface meshing and refinement, were encouraged. The challenge attracted 733 registrations from all continents. In the first phase, 17 teams submitted their algorithms, while in the second phase, 14 teams participated.}
    \label{fig:Challenge organization and participation}
\end{figure}

The challenge achieved global reach, attracting 733 registered users. The geographic distribution was led by participants from Asia (67\%), followed by Europe (29\%) and North America (10\%). Within these regions, China and the United States were the most represented countries, constituting 41\% and 8\% of all participants, respectively. A visual summary of the challenge participation and geographic distribution is provided in \autoref{fig:Challenge organization and participation}.

Phase 1 received 216 algorithm submissions from 17 participating teams, with each team's top-performing submission qualifying for the interim leaderboard ranking. Phase 2 narrowed to 14 teams submitting 21 refined algorithms, maintaining the single-best-submission-per-team selection principle. The three highest-ranking teams advanced to Phase 3 for final algorithm resubmission. Evaluation began in Phases 1 and 2 with standard metrics: the Dice similarity coefficient (DSC) for volumetric overlap and the Hausdorff distance (HD) for surface deviation. The final phase introduced a more rigorous robustness assessment using first- and total-order Sobol' indices. This progressive approach was designed to guide algorithmic refinement and ensure the final ranking accounted for model uncertainty. A complete description of the evaluation metrics is available in \autoref{subsec:challenge evaluation metrics}.

Consistent with prevailing trends in medical image analysis, all submissions implemented deep learning frameworks. The 3D U-Net architecture~\cite{cciccek20163d} emerged as the predominant backbone, adopted by 43\% of submissions. Remaining approaches employed hybrid architectures integrating U-Net variants with complementary network components (convolutional neural network with vision transformer~\cite{dosovitskiy2020image}, U-Net with convolutional block attention module~\cite{woo2018cbam}, residual U-Net~\cite{zhang2018road}, attention U-Net~\cite{oktay2018attentionunetlearninglook}, SegResNet~\cite{myronenko20193d}). Two distinct strategies were observed in the choice of loss functions. Approximately 36\% of the teams relied on standard implementations of the Dice loss, whereas others adopted different approaches, combining multiple objective functions. These included cross-entropy variants (standard or binary), focal loss~\cite{lin2017focal}, Hausdorff distance loss~\cite{karimi2019reducing}, and topology-preserving loss~\cite{clough2020topological}. This dichotomy reflects the field's ongoing exploration of optimization strategies balancing segmentation accuracy with anatomical plausibility.

\begin{figure}[htbp]
    \centering
    \includegraphics[]{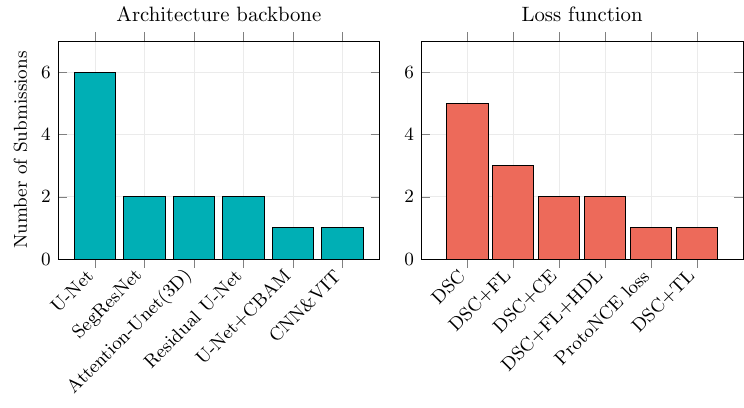}\\
    \caption{Technical overview of leaderboard submissions. All participants adopted deep learning-based approaches to address the challenge task. Most approaches were built upon three-dimensional U-Net architecture, with the Dice loss function serving as the primarily optimization objective. CNN = convolutional neural network, VIT = vision transformer, CBAM = convolutional block attention module, DSC = dice similarity coefficient, CE = cross-entropy, FL = Focal loss, HDL = Hausdorff distance loss. Top: Model architecture backbones. Bottom: Loss functions.}
    \label{fig:technical details}
\end{figure}

The majority of contributions were developed on top of established medical machine learning frameworks, most prominently nnU-Net~\cite{isensee2021nnu} and the Medical Open Network for Artificial Intelligence (MONAI)~\cite{cardoso2022monai}. Furthermore, participants consistently incorporated preprocessing steps, including intensity normalization, spatial resampling, and data augmentation, to enhance segmentation accuracy and robustness. An overview of the employed deep learning architectures and associated loss formulations is presented in \autoref{fig:technical details}.

The top teams in the SEG.A challenge highlighted the critical role of image preprocessing and augmentation in the segmentation task. Intensity normalization and spatial resampling proved particularly effective for handling multi-center datasets. Augmentation techniques -- such as Gaussian noise, intensity scaling, intensity shifting, blurring, and random cropping -- enhanced data utilization, addressing the challenge of limited clinical data. Moreover, the best-performing algorithms demonstrated a preference for fusion mechanisms, enabling the simultaneous segmentation of an aortic vessel tree with varying thicknesses.

\subsection{Segmentation performance}
Methodological details of the highest-ranking submissions are provided in \autoref{sec:methods}.

\begin{figure}[htbp]
\centering
\includegraphics[]{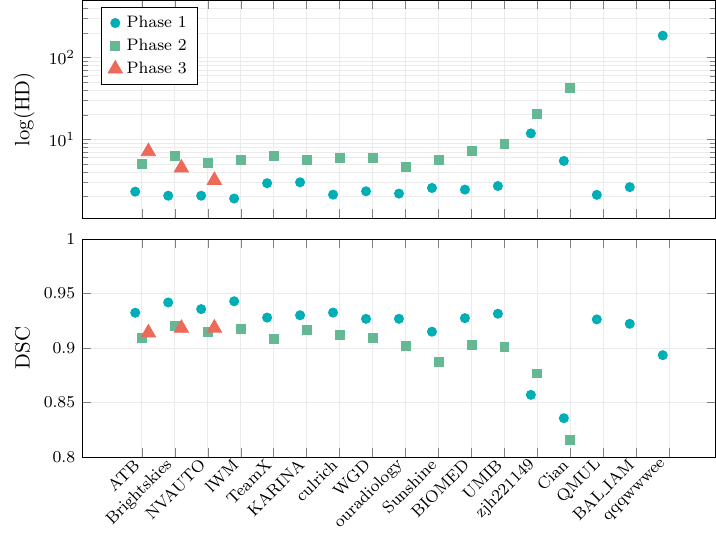}\\
\caption{An overview of algorithm performance is presented, with challenge outcomes evaluated using DSC and HD. Results are ordered from best (left) to worst (right). Each dot represents the mean value of the metric for each of the three challenge phases.}
\label{fig:DSC and HD}
\end{figure}

Overall, the highest-ranking teams exhibited noticeable differences between the two metrics—DSC and HD—highlighting that both are meaningful and provide complementary insights (\autoref{fig:DSC and HD}).

In Phase 1, the mean DSC value produced by each algorithm ranged between $0.836$ and $0.942$, where a value of DSC$=1.0$ represents a full match. Similarly, the mean HD value ranged between $1.9$ and $185.5$, with lower values indicating closer agreement with the reference standard. In Phase 2, the mean DSC ranged between $0.816$ and $0.920$, and the mean HD value ranged between $4.6$ and $42.1$. As noticeable, 
we found that both the DSC and HD results were significantly better in Phase 1 compared to Phase 2. This discrepancy is likely attributable to the extremely limited number of test samples (two) in Phase 1, which was intentionally designed as a warm-up stage.

The top-performing team in Phase 3 resulted to be NVAUTO, who secured the first place in variability interaction ($p_\mathrm{inter}$) and HD metric ($p_\mathrm{HD}$), ranked second in image variability ($p_\mathrm{var}$), and placed third in the DSC metric ($p_\mathrm{DSC}$). The final rankings for Phase 3 are summarized in \autoref{tab:final-ranking}.

\begin{table}[htbp]
  \centering
  \caption{Final Rankings for Phase 3. Here, the symbol $r_\mathrm{X}$ denotes the rank of each corresponding metric $\mathrm{X}$, and NR indicates that a metric is non-ranked due to a bug which resulted in a null output. For more details on the challenge evaluation metrics, please refer to \autoref{subsec:challenge evaluation metrics}.}
  \label{tab:final-ranking}
  \begin{tabular}{lccccccccc|c}
    \toprule
    \textbf{TEAM} & \textbf{$p_\mathrm{DSC}$} & \textbf{$r_\mathrm{DSC}$} & \textbf{$p_\mathrm{HD}$} & \textbf{$r_\mathrm{HD}$} & \textbf{$p_\mathrm{var}$} & \textbf{$r_\mathrm{var}$} & \textbf{$p_\mathrm{inter}$} & \textbf{$r_\mathrm{inter}$} & \textbf{$p_\mathrm{fin}$} & \textbf{$r_\mathrm{fin}$} \\
    \midrule
    NVAUTO      & 2.45 & 3  & 1.15 & 1  & 0.52 & 2  & 1.13  & 1  & 1.83 & 1 \\
    Brightskies & 2.40  & 2  & 2.40  & 2  & 0.51 & 3  & 1.16 & 2  & 2.17 & 2 \\
    ATB        & 1.15 & NR & 2.45 & NR & 0.58 & NR & 1.24  & NR & 3.00 &  3 \\
    \bottomrule
  \end{tabular}
  \vspace{4pt} 
\end{table}

When comparing the three finalists with the twelve best-performing submissions, the results of the remaining contributions across both phases of the challenge were heterogeneous and generally less effective than those of the awarded teams. With few exceptions, the award-winning submissions consistently outperformed the others with respect to both evaluation metrics. For instance, although Team IWM did not advance to Phase 3, it achieved a higher DSC value than ATB and NVAUTO.

\begin{figure}[htbp]
\centering
\includegraphics[width=0.9\textwidth]{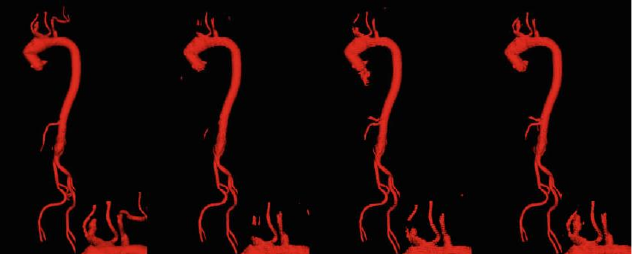}
\caption{Qualitative examples of automated AVT segmentation. From left to right: the ground truth mask, the segmentation produced by Attention U-Net, the segmentation produced by nnU-Net, and the segmentation produced by M3F~\cite{byeon2023m3f}.}
\label{fig:Qualitative examples of automated AVT segmentation}
\end{figure}

Representative qualitative examples of model performance and error cases are shown in \autoref{fig:Qualitative examples of automated AVT segmentation}. In general, false-positive segmentations mainly occurred in regions of thin aortic branches, while aortic segments adjacent to physiological structures were more frequently missed, resulting in false-negative segmentations. These insights underscore the necessity for advanced segmentation techniques that can effectively differentiate between the aorta and adjacent structures, particularly in areas with subtle contrast variations.

\subsection{Surface meshing performance}
\begin{figure}[htbp]
\centering
\includegraphics[]{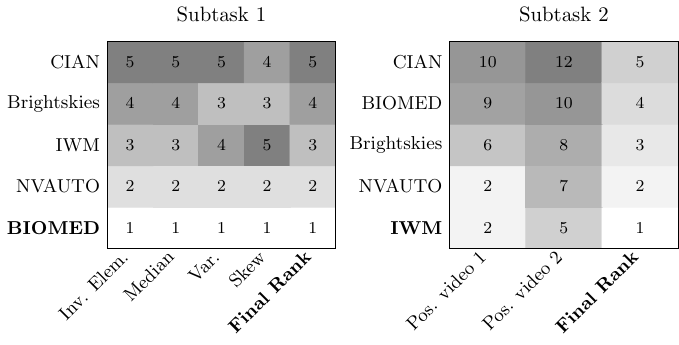}\\
\caption{Final rankings of the two subtasks.}
\label{fig:subtasks}
\end{figure}

Alongside the primary AVT segmentation task, we also evaluated two subtasks: volumetric meshing of the aortic vessel tree and surface meshing for its visualization. \autoref{fig:subtasks} shows the final rankings of the two subtasks of volumetric-ready surface meshing and clinical-grade surface refinement. For each subtask, the rankings of the participating teams across various metrics remain relatively stable. 

Team BIOMED ranked first according to all four evaluation metrics in Subtask 1, namely the median, variance, and skewness of the scaled Jacobian, as well as the average number of invalid elements across all runs. This team used the Marching Cubes algorithm~\cite{lorensen1998marching} to extract a continuous surface mesh from the segmentation mask and then apply improved Laplacian smoothing to enhance its visual quality. Prior to smoothing, the mesh was corrected by removing singularities, self-intersections, and degenerate elements to ensure watertightness. Using \textit{pymeshfix}~(\url{https://pymeshfix.pyvista.org/index.html}), local irregularities are automatically detected and patched by constructing a combinatorial manifold, resulting in a mesh that meets high numerical standards, e.g., the scaled Jacobian.

Team IWM ranked first in Subtask 2, which focused on surface mesh generation for visualization. The team generated surface meshes from the segmentation outputs using the Discrete Marching Cubes algorithm implemented in the \textit{Visualization Toolkit}~(\url{https://vtk.org/}), followed by smoothing with the Windowed Sinc Poly Data Filter algorithm. The chosen parameters included $25$ iterations, a feature angle of $120\textdegree$, a pass band of $0.001$, non-manifold smoothing, and normalized coordinates, while boundary and feature edge smoothing were disabled. For volumetric mesh generation, the parameters were adjusted by enabling boundary and feature edge smoothing and increasing the number of iterations to $30$, while keeping other settings unchanged. Additionally, they closed all mesh holes to ensure watertightness and extended small branches via morphological operations to facilitate successful volumetric mesh generation with \textit{TetGen}~(\url{https://www.wias-berlin.de/software/tetgen}).

\section{Discussion}\label{sec3}
In this summary, we presented the results of SEG.A. challenge on Automated Aortic Vessel Tree (AVT) Segmentation, organized as an official MICCAI satellite event, along with its comprehensive findings and extended analyses to facilitate clinical adoption of automated CTA-based AVT analysis. The challenge focused on establishing benchmark performance for AVT segmentation across multicenter CTA datasets, with top-performing submissions demonstrating that state-of-the-art deep learning architectures can achieve clinically acceptable accuracy for this fundamental clinical task. 

Qualitative analysis revealed critical success factors in algorithm design: sophisticated data-preprocessing strategies emerged as a common feature among leading teams, highlighting the importance of domain-specific knowledge integration alongside systematic error analysis. This observation gains particular relevance given the growing adoption of standardized frameworks like nnU-Net that automate training optimization. Furthermore, we identified a prevalent trend toward multi-branch segmentation architectures in high-performing solutions. These hierarchical networks, which process images at multiple resolutions before fusing features, proved particularly effective for preserving anatomical continuity in vascular structures with heterogeneous lumen diameters, though their computational demands underscore the need for efficiency improvements in future implementations.

Our study also revealed several patient- and image-specific factors that influence algorithmic performance. In particular, segmentation methods often miss smaller vessels. Although these vessels constitute only a tiny fraction of total voxels, so their omission has little effect on aggregate metrics, this limitation underscores the need for more nuanced evaluation measures that explicitly assess fragment detection, instance segmentation, and semantic segmentation in AVT analysis. In clinical contexts such as surgical planning or treatment-response monitoring, relying on a single metric is unlikely to suffice and a tailored combination of complementary quality indicators provides a more accurate appraisal. Ultimately, whether an algorithm’s performance is “good enough” depends on its intended application: for gross aortic geometry quantification , e.g., dissection-risk assessment~\cite{pepe2023Cross}, current approaches may be adequate, whereas the reliable detection of low-contrast, small-caliber segments will require further refinement.

An ensemble of the top-performing algorithms achieved the highest overall performance, consistent with general findings in machine learning, which suggest that combining diverse models often outperforms optimizing a single one. Moreover, our analysis showed that input patch size has a significant impact on performance. Larger patches tend to yield higher segmentation accuracy, but they also lead to larger models that may compromise efficiency and increase operational demands. Future challenges should balance accuracy with usability in clinical settings.

A critical evaluation of the performance metrics revealed a substantial correlation between DSC and HD, highlighing the need for outcome measures that capture complementary aspects of algorithm performance and robustness. To address this, we assessed two additional metrics, the first-order and total-order Sobol' indices~\cite{melito2021sensitivity,Jafarinia2023} during the final evaluation phase. These global sensitivity-analysis metrics quantify the influence that the uncertainty of a model's input variables has on the final output, providing additional insights on a model's robustness towards the intrinsic uncertainty of a multi-center dataset.

The SEG.A. challenge has generated critical insights for advancing automated AVT segmentation. A key insight highlights the need to align quantitative measurements with clinical imaging, especially when developing evaluation frameworks that consider the anatomical differences across the branches of the aorta.

In surgical planning contexts, where accurate delineation of clinically relevant branches supersedes the importance of global segmentation accuracy, contour-based metrics, e.g., HD, demonstrate superior clinical validity compared to conventional volumetric measures~\cite{muller2022towards}. This necessitates a paradigm shift in algorithm development, where loss functions should be explicitly weighted to prioritize branch-specific topological accuracy over aggregate segmentation performance. Furthermore, to address the inherent variability in clinical deployment scenarios -- including heterogeneity in demographics, contrast protocols, and imaging devices --, we advocate integrating global sensitivity analysis, such as Sobol' metrics, during model validation to quantify robustness against these confounding factors. While this iteration established CTA-based segmentation benchmarks, persistent limitations include compromised visualization of low-contrast vascular regions and inadequate representation of pediatric populations, to name a few. These gaps underscore the need for methodological innovations targeting noise-robust feature extraction and age-invariant representation learning in future challenge iterations.

While SEG.A. challenge marks substantial progress in automated AVT segmentation, two persistent challenges still require resolution. While current algorithms demonstrate proficiency in standard imaging conditions, their clinical translation is hindered by: (1) limited generalization across institutions with varying contrast protocols and CT scanner specifications, and (2) insufficient robustness against noise-induced artifacts that disrupt thin vessel continuity. Our future endeavors will therefore prioritize domain-invariant learning frameworks coupled with noise-aware segmentation paradigms, specifically targeting the preservation of topological accuracy in low–signal‐to‐noise vascular regions through hybrid approaches combining physics-based noise modeling and robust feature extraction methods.

\section{Methods}\label{sec:methods}
\subsection{Challenge Mission and Task}
The SEG.A. challenge aims to stimulate research on automated CTA-based AVT segmentation, provide a platform for systematic algorithm comparison, and document the current state of the art in this domain.

The SEG.A. challenge task, which focuses on fully automated segmentation of the AVT, represents a critical step toward quantitative aortic diagnosis, staging, and assessment of therapy response in multicenter CTA datasets. Although manual segmentation is possible, it is highly labor-intensive and thus infrequently performed in clinical practice. Automating AVT segmentation can substantially facilitate clinical adoption and enable comprehensive quantitative analysis. The task is formulated as a semantic segmentation problem rather than an instance segmentation one, owing to the aorta's complex structure: as the largest artery in the body, it comprises multiple segments exhibiting considerable variability in diameter, curvature, and branching patterns~\cite{Pepe2020Detection}. This anatomical complexity poses a significant challenge for accurately delineating and segmenting the entire AVT along with its branches.

We envision the SEG.A. challenge as the inaugural event in a series of initiatives aimed at addressing progressively more complex aspects of automated AVT analysis. Future challenges may encompass the detection and segmentation of individual aortic segments, the application of alternative surface meshing techniques, the use of contrast agents, segment phenotyping, and the evaluation of longitudinal imaging studies.

\subsection{Challenge Organization and Infrastructure}
The SEG.A. challenge was held in 2023 as a registered event of the Medical Imaging Computing and Computed Assisted Intervention (MICCAI) Society. The organizing team was a multidisciplinary group of engineers, radiologists,
and medical data scientists from Graz University of Technology, Austria, and Essen University Hospital, Germany.

The proposal was submitted to MICCAI in December 2022, approved following peer review in February 2023, and is now publicly available~\cite{Pepe2023Towards}. The challenge methodology and results were presented at a Satellite Event during the 26th International MICCAI Conference in October 2023.

The competition officially launched on May 15, 2023, with the release of the public training dataset and all relevant information. The algorithm submission process was structured into three phases. In Phase 1, beginning on June 15, 2023, participants could submit up to six Docker containers for testing on two hidden cases. Phase 2, starting on July 19, 2023, aloowed up to three submissions, evaluated on five hidden cases. The top three solutions progressed to Phase 3, starting on August 16, 2023, where they were tested against 150 augmented test samples. 

The technical execution was managed on the Grand Challenge platform~\footnote{grand-challenge.com, Diagnostic Image Analysis Group, Radboud University Medical Center. The Netherlands}. It was implemented as a type-II challenge, where participants submit their algorithms for execution on a private test dataset. To ensure data confidentiality, submissions were packaged as Docker containers. Participants could modify a provided template and utilize up to five test submissions to validate their implementation on the available computational resources: one TPU (16 GB VRAM) and an 8-core CPU (30 GB RAM).

To aid participants, the organizers provided a baseline algorithm, a complete codebase, and detailed submission instructions, all publicly available through a dedicated online repository~\footnote{\url{https://github.com/apepe91/SEGA2023/tree/main/SegaAlgorithm}} and the challenge website.

\subsection{Participation Policy}
The challenge's top-performing teams were recognized with public announcements and monetary prizes. Award eligibility was contingent on participants submitting either their code in a reproducible format (such as a Docker container) or a technical manuscript describing their methods and results. While teams affiliated with the organizing institutions were allowed to participate, they were excluded from prize consideration. Finally, all participants, regardless of ranking, were invited to co-author this manuscript.

\subsection{Dataset}
\label{subsec:dataset}
The public training data consisted of 56 anonymized, multicenter CTA scans with expert-validated AVT segmentation labels. Sourced from three diverse cohorts: the KiTS19 Grand Challenge~\cite{heller2019kits19}, the Rider Lung CT dataset~\cite{zhao2015coffee}, and Dongyang Hospital. This dataset has been previously described in detail by Radl et al.~\cite{Radl_2022}. The scans provide comprehensive anatomical coverage of the aortic tree, from the ascending aorta to the iliac arteries, and include a variety of pathologies such as aortic dissections and abdominal aortic aneurysms. A detailed overview of the dataset's properties is available in \autoref{tab:databases}. In contrast, the private test dataset and its corresponding labels remained confidential and were accessible only to the organizing committee.

\begin{table}[h!]
\centering
\caption{The detailed overview of the dataset's properties. AD = aortic dissections, AAA = abdominal aortic aneurysms.}
\label{tab:databases}
\begin{tabular}{cccc}
\hline
\textbf{Image Information} & \textbf{KiTS} & \textbf{RIDER} & \textbf{Dongyang} \\
\hline
\textbf{x/y resolution} & 512 $\times$ 512 & 512 $\times$ 512 & 512 $\times$ 666 \\

\textbf{Axial slices} & 94/146/1059 & 260/1008/1140 & 122/149/251 \\

\textbf{Slice thickness} & 0.5/5/5 mm & 0.625/0.625/2.5 mm & 2/3/3 mm \\

\textbf{Pathologies} & None & AD, AAA & None \\

\textbf{Vessel Tree Volume} & 55.8/284.5/464.4 ml & 176.6/354.1/614.0 ml & 126.0/254.9/488.1 ml \\

\textbf{Segmentation Times} & 30/118/30.5 min & 27/422/80 min & 12/35/19.5 min \\

\textbf{Number of Cases} & 20 & 18 & 18 \\
\hline
\end{tabular}
\end{table}

The test dataset comprised 157 whole-body CTA scans with manually segmented AVT labels. This set was composed of two distinct subsets: seven real clinical cases from the same institutions and protocols as the training data, and 150 synthetically generated samples created via data augmentation, as described in \autoref{subsec:Data augmentation}. 

\begin{figure}[htbp]
\centering
\includegraphics[width=0.9\textwidth]{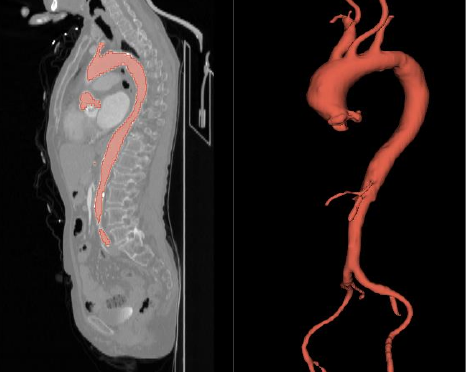}
\label{fig:examples of dataset}
\caption{Example of one CTA image with the AVT mask from the dataset applied in the SEG.A. challenge.}
\end{figure}

All ground truth masks were manually created following these steps. Firstly, the CTA data (.nrrd files) were visualized using 3D Slicer~(\url{https://www.slicer.org/})~\cite{fedorov20123d}, enabling side-by-side and fused views of the medical images. Pre-processing was applied to reduce image noise for segmentation. Gradient anisotropic diffusion, known for its edge-preserving properties, was used, with the parameters listed in \autoref{tab:diffusion_params}. Local thresholding was applied based on manually defined threshold ranges. The GrowCut~\cite{vezhnevets2005growcut} algorithm was then used for coarse segmentation of the aortic vessel tree. Finally, manual post-processing was performed to refine the segmentation and create the final AVT masks (.seg.nrrd file). A representative example of one CTA image with the AVT mask is provided in \autoref{fig:examples of dataset}.

\begin{table}[ht]
\centering
\caption{Parameters for gradient anisotropic diffusion.}
\label{tab:diffusion_params}
\begin{tabular}{lrrr}
\toprule
Case Description & Conductance & Iterations & Time Step \\
\midrule
Little Noise/High Resolution & 0.85 & 1 & 0.0625 \\
Little Noise/Low Resolution  & 0.8  & 1 & 0.0625 \\
Aortic Dissection            & 0.7  & 1 & 0.0625 \\
\bottomrule
\end{tabular}
\end{table}

\subsection{Data augmentation}
\label{subsec:Data augmentation}
Data augmentation is a widely adopted technique in deep learning that expands training datasets through transformations such as intensity adjustment, rotation, noise injection, and filtering. It addresses challenges posed by limited or imbalanced data, which can otherwise lead to overfitting and degraded model performance. By increasing both the size and diversity of the training set, data augmentation improves model generalization and contributes to more accurate predictions.

Data augmentation is particularly valuable in medical imaging, where large-scale datasets are often limited. Simple augmentation approaches can be applied to both 2D and 3D medical images, enabling models to capture critical features beyond those present in the original data. This enhances model robustness and reliability. However, selecting the optimal number of augmented samples and fine-tuning augmentation parameters remain important challenges.

For the evaluation stage of this challenge, data augmentation is embedded within the sensitivity analysis framework to generate and propagate image uncertainty through the submitted algorithms. By introducing controlled variations of the original dataset, augmentation enables the assessment of a model’s sensitivity to input perturbations. This, in turn, provides insights into model reliability and robustness across diverse scenarios, ultimately supporting more consistent and accurate predictions.

\begin{figure}[htbp]
\centering
\includegraphics[]{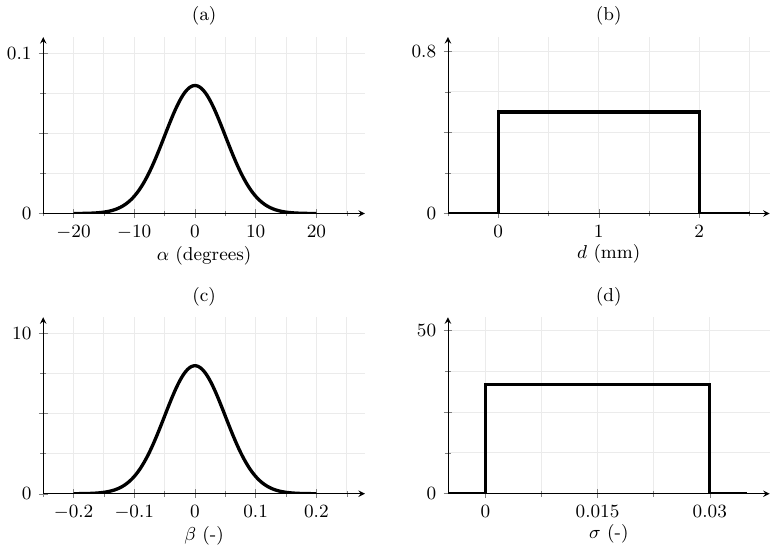}
\label{fig:variablePDFs}
\caption{Probability distributions of the four data augmentation parameters:
a) rotation angle $\alpha$, applied around the z-axis and sampled from a normal distribution with mean $0^\circ$ and standard deviation $5^\circ$;
b) displacement to simulate patient motion, sampled from a uniform distribution between 0 and 2 mm;
c) contrast adjustment using $\gamma = \exp(\beta)$, where $\beta \sim \mathcal{N}(0, 0.05)$;
d) additive Gaussian noise with zero mean, where the standard deviation is sampled from a uniform distribution between 0 and 0.03.}
\end{figure}

Geometric transformations play an essential role in medical imaging by ensuring accurate visualization of anatomical structures. Rotation, a commonly used transformation, is applied to adjust scan orientation. In this simulation, rotations are performed around the z-axis, with angles sampled from a normal distribution centered at $0\textdegree$ and a standard deviation $5\textdegree$ as presented in \ref{fig:variablePDFs}(a). This small variation simulates patient rotation, which may occur due to pathological conditions affecting stability during scanning.

Motion simulation is another critical factor, as patient movement during acquisition can introduce blurring and artifacts that compromise diagnostic accuracy. In the augmentation procedure, motion is modeled using a uniform probability distribution, applying displacements of up to 2 millimeters, a range consistent with clinical observations, as presented in \ref{fig:variablePDFs}(b). The uniform distribution reflects the equal likelihood of movement in any direction, making it suitable for modeling this augmentation property.

Intensity modification is widely employed to adjust image contrast. In this case, contrast alterations are applied to the AVT to simulate variability in contrast fluid distribution. Specifically, gamma correction is used, where $\gamma$ is defined as $\exp(\beta)$, with $\beta$ sampled from a normal distribution with mean zero and standard deviation equal to 0.05, as shown in \ref{fig:variablePDFs}(c).

Finally, noise injection is introduced to enhance model robustness against fluctuations in image quality. Gaussian noise with zero mean is applied, with the standard deviation sampled from a uniform distribution between $0$ to $0.03$, as presented in \ref{fig:variablePDFs}(d). This augmentation accounts for diverse noise sources, including electronic noise, detector artifacts, and patient motion, thereby improving the model's resilience under real-world imaging conditions.

\subsection{Challenge evaluation metrics}
\label{subsec:challenge evaluation metrics}
Four quantities were used to define the evaluation metrics and assess the submitted algorithms: the Dice Similarity Coefficient (DSC), Hausdorff Distance (HD), and the first-order and total-order Sobol' indices. In the first phase of the challenge, evaluation was based on the mean, variance, and interquartile range of DSC and HD computed from the submitted algorithms. In the second phase, two additional metrics were introduced to account for the distribution of DSC and HD across all simulated hidden cases. 

The DSE-related metric was defined as a weighted aggregation of the generated rankings. Ideally, DSC values should exhibit minimal variation and remain as close to one as possible. However, given the skewed distribution of DSC, we evaluated the median ($m_\mathrm{DSC}$), variance ($\sigma^2_\mathrm{DSC}$), and skewness ($\mu_{3,\mathrm{DSC}}$) for each submitted algorithm. This comprehensive evaluation captured different performance characteristics, including central tendency, dispersion, and asymmetry. Accordingly, the final metric $p_\mathrm{DSC}$ is defined as
\begin{equation}
    p_\mathrm{DSC} = 0.6\times r_{m_\mathrm{DSC}}+0.25\times r_{\sigma^2_\mathrm{DSC}}+ 0.15\times r_{\mu_{3,\mathrm{DSC}}}.
\label{equ:dsc}
\end{equation}

The HD-related evaluation metric is ideally as close to zero as possible, with minimal variation. Similarly, the evaluation methodology for HD employs the calculated rankings of the median ($m_\mathrm{HD}$), variance ($\sigma^2_\mathrm{HD}$), and skewness ($\mu_{3,\mathrm{HD}}$) of this loss function. The resulting rankings were then combined using the same weighting scheme applied to DSC, yielding the second evaluation metric $p_\mathrm{HD}$ as
\begin{equation}
    p_\mathrm{HD} = 0.6\times r_{m_\mathrm{HD}}+0.25\times r_{\sigma^2_\mathrm{HD}}+ 0.15\times r_{\mu_{3,\mathrm{HD}}}.
\label{equ:hd}
\end{equation}

In the final phase, we introduced two additional metrics to quantify the robustness of the submitted algorithms to the image variations imposed in the augmentation phase:  $p_\mathrm{var}$ and $p_\mathrm{inter}$. 
While the first assesses how unbalanced the algorithm is with respect to the four different image variations, the latter indicates how sensitive the outputs produced by the submitted algorithm, i.e., DSC and HD, are to their interactions.

These metrics used the first-order $S_i^{1}$ and total-order Sobol' indices $S_i^{\mathrm{T}}$, which quantify the impact of image variation, such as intensity, rotation, translation, noise, and blur, on the metrics DSC and HD. The first-order Sobol' index measured the individual influence of each input variable, while the total-order Sobol' index also captured the effect of their interactions on the final outcome. The difference between $S_i^{\mathrm{T}}$ and $S_i^{\mathrm{1}}$ isolates the interaction effect of the variables on the algorithm.

Ideally, the metrics DSC and HD should be equally sensitive to variations across all input images. If this condition holds, the algorithm can be regarded as robust and uniformly responsive to different sources of image variability. Conversely, if either metric exhibits disproportionate sensitivity to specific input factors, it suggests that the model may be overly influenced by variations along those particular dimensions.

Similarly, it is desirable that the algorithm's input variables exhibit minimal or no interaction effects. This condition facilitates the independent assessment of each variable’s contribution to model performance, which is critical for targeted improvements. Moreover, the absence of interactions implies that issues can be addressed within specific input domains without unintended consequences in others, thereby supporting more controlled and interpretable model refinement.

The metrics $p_\mathrm{var}$ and $p_\mathrm{inter}$ are finally defined as
\begin{subequations}
\begin{align}
    p_\mathrm{var} = 1-\sum^{M}_{i=1}\left|S_i^{1}-\frac{1}{M}\right|, \label{equ:variability}\\
    p_\mathrm{inter} = \sum^{M}_{i=1}(S_i^{\mathrm{T}}-S_i^{1}), \label{equ:interaction}
\end{align}
\end{subequations}
where $M$ is the total number of image variations, which is four in this study. As equation~\ref{equ:variability} approaches the unity, $p_\mathrm{var}$ indicates a uniform distribution of the first-order Sobol' indices (the ideal case), while a $p_\mathrm{inter}$ value closer to 0 indicates less interaction between the variables, as measured by the algorithm.

Volume-based metrics were selected for their clinical relevance, as many key biomarkers -- such as aortic volume and total lesion glycolysis -- are volume-dependent. Additionally, accurate volume-based segmentation is essential for applications such as local treatment planning (e.g., surgical interventions) and the assessment of therapy response.

An intermediate ranking was generated for each evaluation metric, namely $r_\mathrm{DSC}$, $r_\mathrm{HD}$, $r_\mathrm{var}$, and $r_\mathrm{inter}$. The final ranking was then derived by computing their weighted average
\begin{equation}
    r_\mathrm{fin}=\frac{(r_\mathrm{DSC}+r_\mathrm{HD})}{6}+\frac{(r_\mathrm{var}+r_\mathrm{inter})}{3}.
\label{equ:finalRanking}
\end{equation}
In the event of tied rankings, the algorithm with the lowest average execution time was given the higher rank.

\subsection{Subtasks metrics and analyses}
Subtask 1 of the AVT volumetric surface meshing was evaluated using the Jacobian-based methodology, following a procedure similar to that used for the main metrics. The evaluation considered assessing the median ($m_{J}$), variance ($\sigma^2_{J}$), and skewness ($\mu_{3,J}$) of the scaled Jacobian distribution for each geometry. Additionally, the average number of invalid elements -- defined by a negative scaled Jacobian -- was computed across all runs and is denoted as $n$. These metrics were then ranked accordingly. The final metric, $p_J$, was a weighted combination of these rankings, with weights of 0.3, 0.25, 0.15, and 0.3, to capture various performance aspects
\begin{equation}
    p_J = 0.3\times r_{m_{J}}+0.25\times r_{\sigma^2_{J}}+ 0.15\times r_{\mu_{3,J}}+0.3\times r_n.
\label{equ:subtask1}
\end{equation}

Subtask 2, which involved clinical-grade surface refinement, was evaluated by experienced medical professionals using a Likert-scale questionnaire. The qualitative assessment emphasized the number of relevant branches and the absence of segmentation artifacts. Eight professionals from five medical institutions across four countries participated: Austria (three), Germany (three), Iran (one), and the UK (one). The participants included board-certified vascular radiologists (three), cardiac surgeons (three) and senior research clinicians specializing in vascular imaging (two). All participants indicated that their work would at least partially benefit from the automated workflow. The evaluation questions, which pertained to the outputs of each team, are listed in \autoref{tab:questions}.

\begin{table}[htbp]
    \centering
    \caption{Questions answered by the medical professionals. All questions could be answered using a 7-point Likert scale.}
    \label{tab:questions}
    \begin{tabularx}{\textwidth}{lX}
    \toprule
    \# & Question  \\
    \midrule
    1. & Does the 3D model communicate all relevant information for clinical needs? \\
    2. & Is the 3D model free from artifacts (e.g. voxel artifacts)? \\
   \\
    \bottomrule
    \end{tabularx}
    \vspace{4pt} 
\end{table}

\subsection{Highest-ranking submissions}
Below, we present concise descriptions of the eight top-ranked submissions based on the final leaderboard, followed by comprehensive evaluations for all submissions. The analysis results were derived from executing the submitted algorithms on the test dataset and were consistent with the official leaderboard rankings. Reproducibility of all algorithms is ensured through the provision of Docker containers.

\subsubsection{Team ATB}
Team ATB introduced the Multi-Field-of-View Feature Fusion Network (M3F) based on an attention U-Net architecture for AVT segmentation. The network employs a dual-branch approach to effectively fuse contextual information from broad patterns with fine details from high-resolution patches, significantly enhancing segmentation accuracy~\cite{byeon2023m3f}.

\subsubsection{Team NVAUTO}
Team NVAUTO's winning solution built upon the Auto3DSeg framework from MONAI, employing an ensemble of 15 SegResNet models to achieve robust aorta segmentation performance. Their approach integrated multiple deep learning architectures and advanced optimization strategies~\cite{myronenko2023aorta}.

\subsubsection{Team BRIGHTSKIES}
Team BRIGHTSKIES demonstrated that a data-centric approach, combined with a chain of U-Net networks, enabled accurate AVT segmentation. They showed that preprocessing steps such as histogram matching and sigmoid windowing enhanced model generalization across varying imaging conditions and improved robustness to noisy data. The incorporation of topological loss in the refining stage improved the connectivity of segmented structures, while the interpolation refinement model generated smoother and more realistic outputs~\cite{el2023data}.

\subsubsection{Team IWM}
Team IWM, winner of subtask 2 -- visualization, concluded that their heavily augmented, high-resolution 3-D ResUNet approach provided a robust and stable solution for automatic aorta segmentation. The team demonstrated that comprehensive data preprocessing and aggressive augmentation were crucial in overcoming the challenges posed by limited training data and heterogeneous imaging conditions~\cite{wodzinski2023automatic}.

\subsubsection{Team KARINA}
Team KARINA developed a two-stage segmentation framework that combined coarse segmentation with a prototype-based fine segmentation approach. Their method incorporated position-encoded pixel-to-prototype contrastive learning to capture nuanced aortic features~\cite{kim2023position}. 

\subsubsection{Team Qmul}
Team Qmul introduced a novel Misclassification Loss function that leveraged a differentiable XOR operation to identify misclassified voxels and fine-tune a pre-trained segmentation network. Their approach effectively reduced both false positives and false negatives, leading to improved segmentation accuracy compared to conventional loss functions~\cite{khan2023misclassification}.

\subsubsection{Team BIOMED}
Team BIOMED, winner of subtask 1 -- volumetric meshing, demonstrated that their SegResNet-based methodology effectively performed AVT segmentation from CTA images while also providing high-quality mesh reconstructions. Their approach integrated rigorous preprocessing, data augmentation, and an advanced mesh refinement pipeline to produce watertight surfaces suitable for clinical applications~\cite{vagenas2023deep}. 

\subsubsection{Team FZU-IIPALab}
Team FZU-IIPALab developed RASNet, a robust aortic segmentation network built on a 3D U-Net framework enhanced with convolutional block attention modules. Their approach incorporated an adaptive HU preprocessing strategy to mitigate intensity variability across multicenter datasets and employed a two-stage training process that introduced Hausdorff distance loss to refine the segmentation structure~\cite{zhang2023rasnet}. 

\section{Data availability}
As mentioned in \autoref{subsec:dataset}, the training data are publicly available~\cite{Radl2022avtData}. The test data are withheld from public release because they constitute the private test set designated for use in future iterations of the SEG.A. challenge. Use of the private test data for analyses is permitted only following the completion of the SEG.A. challenge series and provided that such use does not interfere with the execution of the challenge. Data requests may be directed to the organizing team via \url{https://multicenteraorta.grand-challenge.org/organizing-team/}.

\section{Code availability}
All code supporting data processing and performance analysis is openly accessible on GitHub at \url{https://github.com/apepe91/SEGA2023/tree/main/SegaAlgorithm} and is distributed under the MIT license.




\bibliography{refs}

\end{document}